\documentclass[10pt,twocolumn,letterpaper]{article}

\usepackage{graphicx}
\usepackage{subcaption}
\usepackage{float}
\usepackage[justification=raggedright]{caption}	
\usepackage{lscape}                                         
\usepackage{wrapfig}

\usepackage[lined,ruled,linesnumbered]{algorithm2e}
\usepackage{animate} 

\usepackage{booktabs}                   
\usepackage{multirow}

\usepackage{makecell}

\usepackage{paralist}
\usepackage{enumitem}
\setlist[enumerate]{itemsep=0mm}

\usepackage{bm}                          
\usepackage{epsfig}                      
\usepackage{graphicx}                  
\usepackage{times}
\usepackage{mathptmx}
\usepackage{mathtools}
\usepackage{amssymb,amsmath}   

\usepackage{units}
\usepackage{color}

\usepackage{comment}

\usepackage{url}  
\usepackage[pagebackref,breaklinks,colorlinks,linkcolor=blue,citecolor=blue,urlcolor=blue,bookmarks=false]{hyperref}
\usepackage{xspace}
\usepackage[table]{xcolor}
\usepackage{setspace}
\usepackage{grfext}
\PrependGraphicsExtensions*{.jpg,.png,.PNG}

\usepackage{gensymb}

\usepackage{soul}
\usepackage{svg}

\usepackage[accsupp]{axessibility} 

\DeclareMathAlphabet{\altmathcal}{OMS}{cmsy}{m}{n}
\DeclareMathAlphabet{\mathbfit}{OT1}{ptm}{bx}{it}

\newlength\paramargin
\newlength\figmargin

\newlength\secmargin
\newlength\figcapmargin
\newlength\tabcapmargin

\newcommand{\mpage}[2]
{
\begin{minipage}{#1\linewidth}\centering
#2
\end{minipage}
}

\newcommand{\mfigure}[2]
{
\includegraphics[width=#1\linewidth]{#2}
}

\newcommand{\topic}[1]
{
\vspace{1mm}\noindent\textbf{#1}
}

\long\def\ignorethis#1{}

\newbox\jsavebox%

\makeatletter
\newcommand{\providelength}[1]{%
  \@ifundefined{\expandafter\@gobble\string#1}
   {
    \typeout{\string\providelength: making new length \string#1}%
    \newlength{#1}%
   }
   {
    \sdaau@checkforlength{#1}%
   }%
}

\newcommand{\sdaau@checkforlength}[1]{%
  \edef\sdaau@temp{\expandafter\sdaau@getfive\meaning#1TTTTT$}%
  \ifx\sdaau@temp\sdaau@skipstring
    \typeout{\string\providelength: \string#1 already a length}%
  \else
    \@latex@error
      {\string#1 illegal in \string\providelength}
      {\string#1 is defined, but not with \string\newlength}%
  \fi
}
\def\sdaau@getfive#1#2#3#4#5#6${#1#2#3#4#5}
\edef\sdaau@skipstring{\string\skip}
\makeatother

\usepackage[capitalize]{cleveref}
\crefname{section}{Sec.}{Secs.}
\Crefname{section}{Section}{Sections}
\Crefname{table}{Table}{Tables}
\crefname{table}{Tab.}{Tabs.}

\def\xi{\mathbf{x}_i}

\graphicspath{{figures}, {examples}}

\makeatletter

\def\@fnsymbol#1{\ensuremath{\ifcase#1\or \dagger\or \ddagger\or
\mathsection\or \mathparagraph\or \|\or **\or \dagger\dagger
\or \ddagger\ddagger \else\@ctrerr\fi}}
\makeatother

\graphicspath{{figures/}}

\usepackage{iccv}

\usepackage{tikz}
\usepackage{pgfplots}
\pgfplotsset{compat=1.13}
\usepackage{pgfplotstable}
\pgfplotsset{every axis/.append style={
grid style=dashed,
}}
\usepackage{subcaption}

\iccvfinalcopy 

\def\iccvPaperID{7066} 

\begin{document}

\title{
3D Motion Magnification: \\
Visualizing Subtle Motions with Time-Varying Radiance Fields
}

\author{Brandon Y. Feng$^*$ \\
University of Maryland\\
\and
Hadi Alzayer$^*$\\
University of Maryland\\
\and
Michael Rubinstein\\
Google Research\\
\and
William T. Freeman\\
Google Research, MIT\\
\vspace{-10pt}
\and
Jia-Bin Huang\\
University of Maryland\\
\vspace{-10pt}
\and
\url{https://3d-motion-magnification.github.io/}
}

\newcommand\blfootnote[1]{%
  \begingroup
  \renewcommand\thefootnote{}\footnote{#1}%
  \addtocounter{footnote}{-1}%
  \endgroup
}
\twocolumn[{
\renewcommand\twocolumn[1][]{#1}
\maketitle
\centering
\vspace{-20pt}
\includegraphics[height=0.36\textwidth]{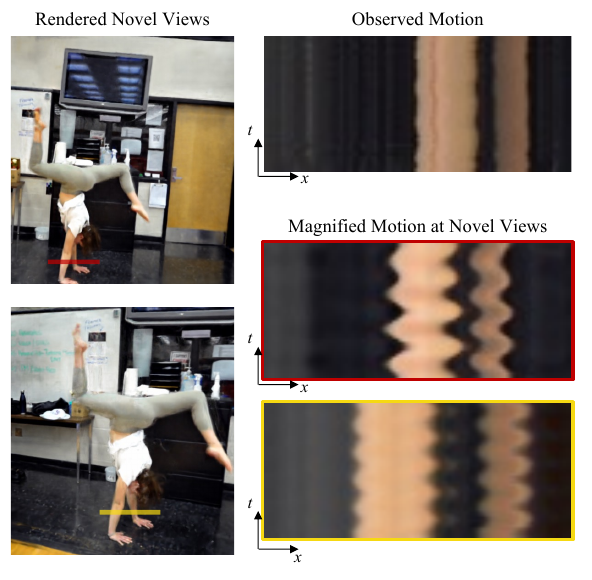} \quad \quad
\includegraphics[height=0.36\textwidth]{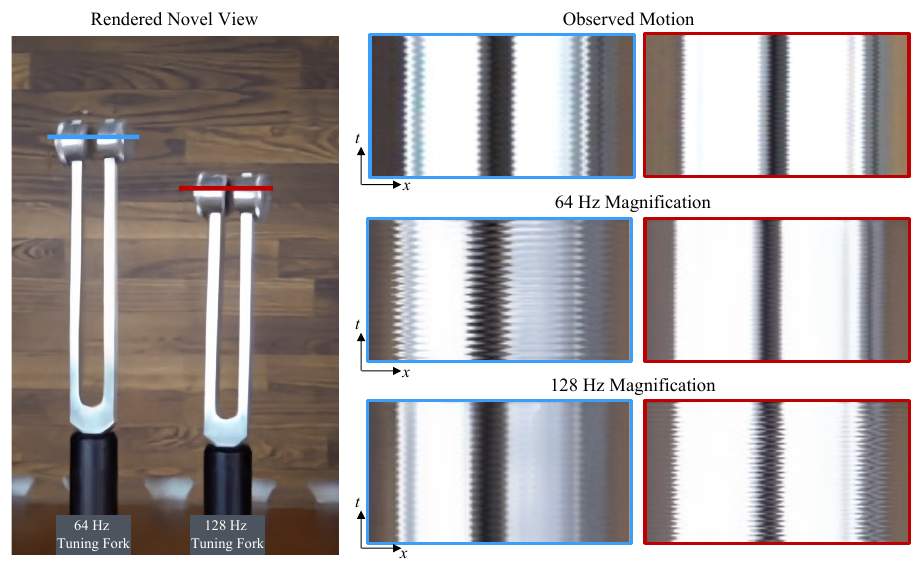}
\vspace{-.2in}
\mpage{0.41}{(a) View synthesis with motion magnification} \hfill
\mpage{0.57}{(b) Frequency-selective magnification with temporal filtering}
\vspace{.1in}
\captionof{figure}{\textbf{3D motion magnification}.
(a) Novel view synthesis with a gymnast doing a handstand while magnifying the small movements of the arms needed to remain balanced. 
(b) Motion magnification based on targeted frequencies through temporal filtering, where the left tuning fork vibrates at 64Hz and the right at 128Hz. 
We visualize x-t (space-time) slices to demonstrate the motion.
\vspace{10pt}}
\label{fig:teaser}



}]


\begin{abstract}
Motion magnification helps us visualize subtle, imperceptible motion. 
However, prior methods only work for 2D videos captured with a fixed camera.
We present a 3D motion magnification method that can magnify subtle motions from scenes captured by a moving camera, while supporting novel view rendering.
We represent the scene with time-varying radiance fields and leverage the Eulerian principle for motion magnification to extract and amplify the variation of the embedding of a fixed point over time.
We study and validate our proposed principle for 3D motion magnification using both implicit and tri-plane-based radiance fields as our underlying 3D scene representation.
We evaluate the effectiveness of our method on both synthetic and real-world scenes captured under various camera setups.
\end{abstract}

\section{Introduction}
\label{sec:intro}
\blfootnote{*Equal contribution}
We live in a big world of small motions.
These motions, such as human respiration or object vibration, are hard to perceive with our naked eyes.
Video processing techniques~\cite{Liu2005MotionM, Wu12Eulerian, Wadhwa2013PhaseBased} have been developed to extract and magnify subtle motions captured in a 2D video to highlight and visualize those motions.
These motion magnification techniques empower visual analytics tools like detecting the vibrations of buildings and measuring a person's heart rate using only a video, without the need for physical contact~\cite{Wang2015ExploitingSR, chen2017video, shang2018multi_civil, lado2023learning_civil}.
\begin{figure}[!t]
    \centering
    \mfigure{0.48}{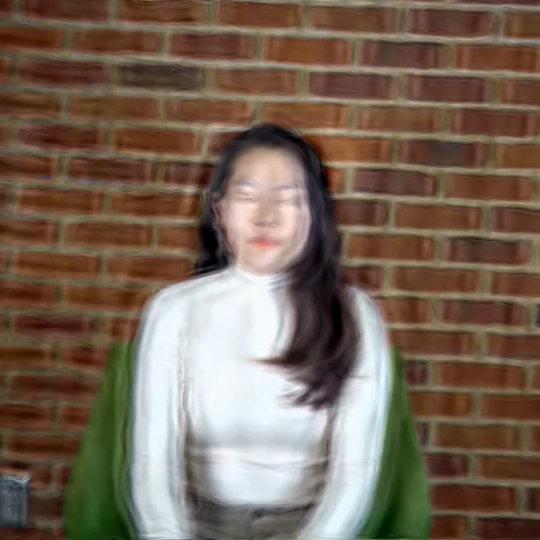} \hfill
    \mfigure{0.48}{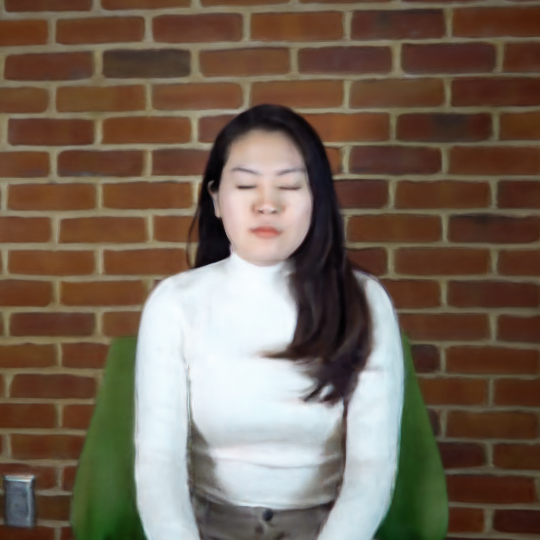}
    \mpage{0.48}{2D Method \cite{wadhwa2014phase}\vspace{2pt}} \hfill 
    \mpage{0.48}{Ours\vspace{2pt}} \\
    \vspace{2pt}
\caption{
\textbf{Motion magnification from a handheld video.} 
Prior 2D motion magnification approaches (e.g.~\cite{wadhwa2014phase}) cannot handle videos captured by a moving camera, producing severe artifacts. 
In contrast, our approach can naturally separate camera motion from object motion, allowing us to magnify only the motion of the subject of interest. 
See Figure \ref{fig:realresult} for our magnified output.}
    \label{fig:handheld}
\end{figure}

\begin{figure*}[!t]
    \centering \includegraphics[width=0.98\linewidth]{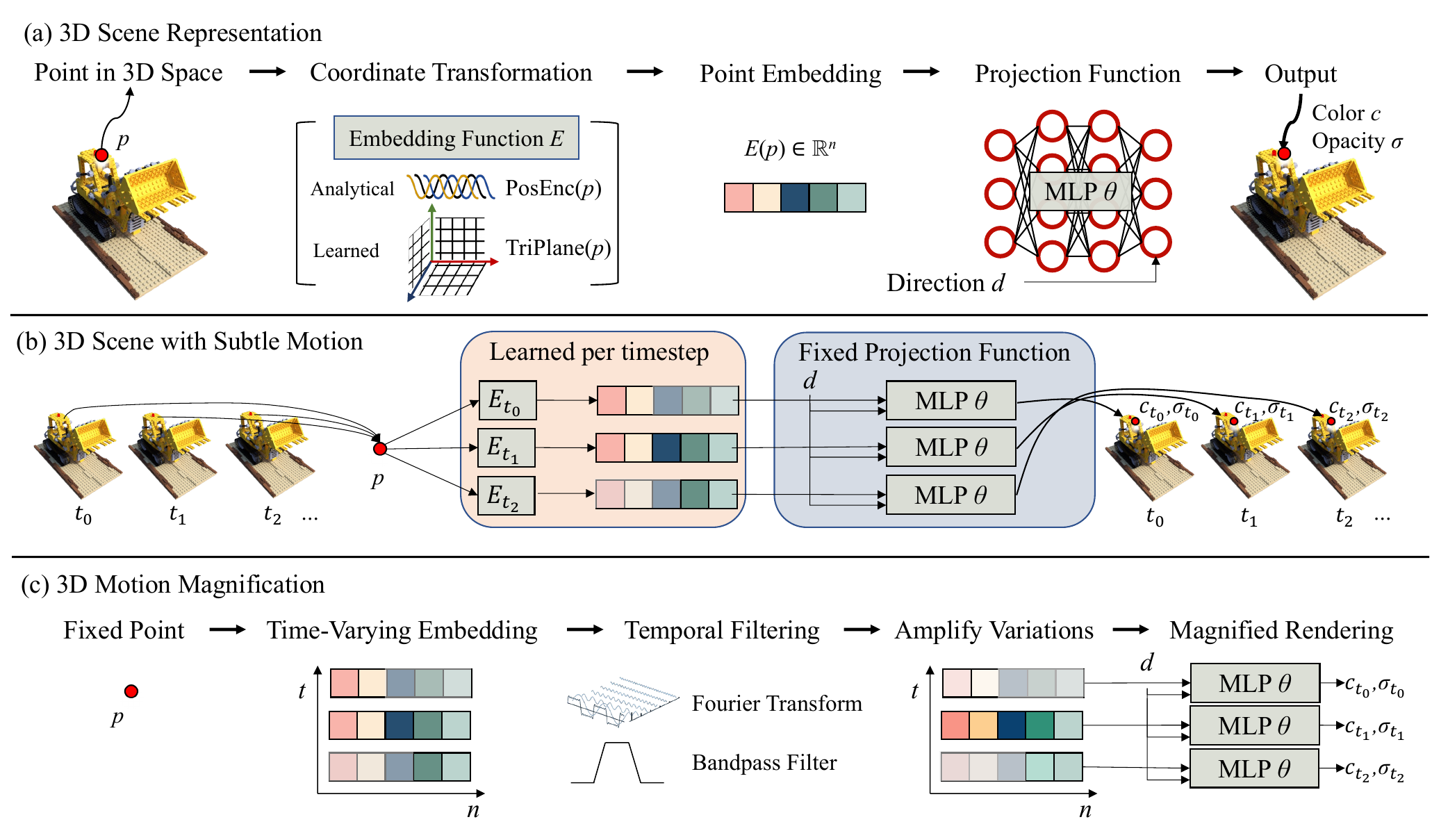}
\caption{\textbf{Method overview.} 
(a) 3D scene representation with NeRF consists of two main components: 
1) Coordinate Transformation uses an embedding function $E$ to map the input point $p\in \mathbb{R}^{3}$ to a high-dimensional embedding vector $E(p)\in \mathbb{R}^{n}$. 
The embedding function can be analytical (positional encoding) or learned (tri-plane).  
2) The Projection Function $\theta$ (usually an MLP) takes in the point embedding and viewing direction, and regresses them into the output color $c$ and opacity $\sigma$ at $p$.
(b) We study scenes with \textbf{subtle motions}. 
To model the tiny variations with NeRF, we change $E$ over time while fixing the projection function.
(c) At a given point $p$, we analyze its embedding variations over time: $[E_{t_0}(p), ..., E_{t_{T-1}}(p)]$. 
We perform temporal filtering to isolate and amplify embedding variations within a certain frequency range and then send the amplified embedding to the MLP $\theta$, resulting in motion-magnified 3D rendering.}
\end{figure*}

However, we live in a \emph{3D world} full of \emph{3D motions}. 
Magnifying motion in 3D, as shown in Figure \ref{fig:teaser} allows us to perceive these motions from different views. 
Furthermore, modeling the motion in 3D  provides a natural separation between camera motion and the motion of subjects of interest. 
This enables magnifying the motion from \emph{handheld videos}, as shown in Figure~\ref{fig:handheld}. 
In contrast, prior 2D motion magnification methods catastrophically fail in such scenarios.

In this paper, we propose a method for \emph{3D motion magnification} using neural radiance fields (NeRF), with minimal modifications to standard NeRF backbones and training pipelines.
Prior methods designed for 2D videos often leverage the {\it Eulerian} perspective, which analyzes and amplifies the color variations at each pixel location over time to magnify motion.
In contrast, we bring the Eulerian analysis to a new domain beyond color space by designing a magnification method operating on the \emph{feature embeddings} of NeRF.
Our experimental results demonstrate that amplifying temporal variations in the \emph{feature embedding of each 3D point} is highly effective in magnifying subtle 3D motion. 
We observe that magnifying the point embedding provides more accurate and robust magnified renderings than Eulerian magnification performed directly on rendered images.

Using images captured during a time window when \textit{only subtle motion is visible}, we train NeRF to reconstruct the 3D scene with such subtle temporal variations. 
We ensure that the only element that changes {\it over time} is the point embedding function, while the MLP layers of NeRF {\it remain constant over time}. 
Although the linear Eulerian approach~\cite{Wu12Eulerian} is agnostic to data dimensionality and is extensible to point embeddings of NeRF, for the phase-based Eulerian approach~\cite{Wadhwa2013PhaseBased}, which showed superior properties over the linear approach, it remains unclear how it may be applied for NeRF as it specifically constructs a complex steerable pyramid over each 2D image frame.
The recently introduced tri-plane representation for NeRF's embedding function naturally allows for 2D-specific magnification methods like the phase-based approach~\cite{chan2022efficient}.
Instead of using the analytical {\it positional encoding} to generate point embeddings, we learn one feature tri-plane at each observed timestep.
These tri-planes can be naturally organized as feature videos for 2D video-based magnification methods. 
Finally, the motion-magnified 3D scene is rendered using these motion-magnified feature triplanes as the point embedding functions.

To evaluate the performance of 3D magnification with NeRF, we first create a synthetic dataset of scenes with subtle motions and measure the magnification quality against synthetically magnified ground truth videos.
The phase-based approach operating on tri-plane features leads to the best performance compared to other alternative approaches considered in our experiments.
To further validate the practicality of the proposed method, we use our pipeline to process several real-world captured scenes with varying camera setups, scene compositions, and subject motions.
Our results show that our proposed approach for 3D motion magnification achieves robust performance for real-world captures in the presence of image noise and camera poses.

To summarize, our contributions are:
\begin{itemize}
    \item We introduce the problem of 3D motion magnification. We demonstrate the feasibility of applying Eulerian motion analysis for 3D motion magnification using standard NeRF backbones and training pipelines.
    \item We extend Eulerian analysis to a new domain beyond color space, exploring strategies to modify and filter point embedding and comparing their trade-offs.
    \item We demonstrate successful 3D motion magnification results on various real-world scenes with different motions, scene compositions, and even handheld videos unsupported by previous 2D methods.
\end{itemize}
\section{Related Work}
\label{sec:related}

\topic{Video motion magnification.} 
Prior approaches to video magnification fall under two categories, inspired by fluid dynamics: Lagrangian \cite{Liu2005MotionM} and Eulerian \cite{Wu12Eulerian, Wadhwa2013PhaseBased, wadhwa2014phase, oh2018learning, zhang2017}. 
The Lagrangian perspective tracks individual pixels as fluid particles and estimates their motion vectors to warp pixels in the image.
Lagrangian-based approach to motion magnification computes the optical flow \emph{explicitly} and uses the estimated flow to magnify the motions of the pixels \cite{Liu2005MotionM}.
The performance, however, is limited by the accuracy of flow estimation.
On the other hand, the Eulerian perspective analyzes the changes at fixed pixel locations, amplifying the temporal variations at each pixel/location to magnify motion.
This approach bypasses the need for explicit feature tracking or optical flow estimation, which can be inaccurate and costly. 
Two variants of the Eulerian approach are linear~\cite{Wu12Eulerian} and phase-based~\cite{Wadhwa2013PhaseBased,wadhwa2014phase}.
Linear Eulerian~\cite{Wu12Eulerian} constructs Laplacian pyramids over the video frames and amplifies the color variation of each pixel over time.
Phase-based Eulerian~\cite{Wadhwa2013PhaseBased, wadhwa2014phase} operates on the phase variations at each pixel, extracted from a complex steerable pyramid~\cite{freeman1995steerable, freeman91design} decomposition of each video frame.
Later work focuses on magnifying larger motion with affine transform and isolated regions of interest with matting \cite{elgharib15DVMAG}, using linear-based methods instead of hand-designed filters \cite{oh2018learning}, and adopting a second-order approximation (with acceleration) instead of first-order methods~\cite{zhang2017}.
Video-based motion magnification has also been applied to extract signals like sound waves from videos recording objects, like a bag of chips, deform and oscillate~\cite{Davis2014visualmic, Sheinin:2022:Vibration}.
Our work builds upon classical Eulerian motion magnification but extends it 1) from 2D to 3D and 2) from color space to the point embedding space of radiance fields. 
Our results show that the Eulerian principle still holds in the point embedding space.

\topic{Static radiance fields.}
NeRF \cite{mildenhall2020nerf} has become the mainstream approach for representing 3D scenes and demonstrates high-quality view synthesis results.
Various techniques have been introduced to improve NeRF in several aspects, including 
training and rendering acceleration \cite{li2022nerfacc, mueller2022instant,attal2022learning, sfk_kplanes_2023, yu_and_fridovichkeil2021plenoxels,Chen2022ECCV}, reducing aliasing~\cite{barron2021mipnerf}, unbounded scene modeling~\cite{barron2022mipnerf360,zhang2020nerf++}, and optimizing poses~\cite{lin2021barf,meuleman2023progressively}.
Factor Fields~\cite{Chen2023factorfields} present a unified framework summarizing various NeRF variants and other neural signal representations as mainly composed of two components: 
(1) a \emph{Coordinate Transformation} that maps input coordinates into an embedding space, and 
(2) a \emph{Projection Function} that maps the embeddings into a value in the field. 
In this paper, we adopt a similar perspective and focus on analyzing the relationship between point embedding and subtle motions. 
We propose magnifying subtle motions through Eulerian magnifications of point embeddings in NeRF. 
We demonstrate successful applications of this approach on NeRF with both positional encoding-based embedding \cite{mildenhall2020nerf, li2022nerfacc} and tri-plane embedding \cite{chan2022efficient, sfk_kplanes_2023}.

\topic{Dynamic scene representations.}
Extensive research has been devoted to extending NeRF for modeling dynamic scenes.
One line of work learns a deformation field and uses it to warp a canonical NeRF for each timestep~\cite{pumarola2020d,park2021nerfies,park2021hypernerf,tretschk2021non,li2021neuralb}. 
Alternatively, one can directly learn a space-time radiance field with time as an additional coordinate~\cite{li2021neural,xian2021space,gao2021dynamic,attal2023hyperreel,song2022nerfplayer,liu2023robust,sfk_kplanes_2023,cao2023hexplane,shao2023tensor4d}.
A major challenge of reconstructing dynamic scenes is capturing time-synchronous multi-view observations. 
While a multi-camera setting is ideal for acquiring high-quality data \cite{Zhao_2022_CVPR,cai2022humman,Liu2021neuralcator}, researchers have explored the more challenging but practical setting of single-camera captures, leveraging priors such as consistent depth~\cite{xian2021space,luo2020consistent,kopf2021robust}, optical flow~\cite{li2021neural,gao2021dynamic,liu2023robust}, or human prior~\cite{neural-human-radiance-field, weng2022humannerf}. 
We demonstrate the applicability of 3D motion magnification in both multi-camera and single-camera setups.

\topic{Implicit representations.}
Implicit representations have emerged as powerful tools for modeling signals~\cite{sitzmann2020implicit,fathony2020multiplicative,park2019deepsdf,lindell2021bacon, saragadam2022miner,tiwary2023orca,tiwary2022towards,feng2021signet,feng2022prif,feng2022viinter,shekarforoush2022residual}.
Mai and Liu \cite{Mai_2022_PhaseNIVR} study \emph{2D videos} with implicit neural representation and model motions in videos by learning \textit{spatially invariant} phase shifts in the positional encoding function.
Our experiment on magnification based on position encoding functions shares similar ideas. 
However, unlike Mai and Liu~\cite{Mai_2022_PhaseNIVR}, we learn \emph{spatially varying} phase shifts and magnify \emph{3D motion}.

\def\D{\altmathcal{D}}
\def\I{\altmathcal{I}}
\def\O{\altmathcal{O}}
\def\res{\altmathcal{R}}

\def\b{\mathbfit{b}}
\def\c{\mathbfit{c}}
\def\d{\mathbfit{d}}
\def\o{\mathbfit{o}}
\def\p{\mathbfit{p}}
\def\t{\mathbfit{t}}
\def\x{\mathbfit{x}}
\def\z{\mathbfit{z}}

\def\K{\mathbfit{K}}
\def\R{\mathbfit{R}}

\def\ang{\phi}
\def\dehom{\mu}
\def\proj{\pi}
\def\sigmoid{S}
\def\vis{\nu}
\def\r{\mathbfit{r}}

\def\bp{(\p\!)} 
\def\bt{(t\!)} 
\def\bx{(\x\neg)} 

\def\ok{\o_{\neg k}}
\def\tk{\t_{\neg k}}
\def\wk{w_{\neg k}}
\def\xi{\x_{\neg i}}
\def\zk{\z_{\neg k}}
\def\Kk{\K_{\neg k}}
\def\Rk{\R_{\neg k}}

\def\ng{\hspace{-0.1mm}}
\def\neg{\hspace{-0.2mm}}
\def\pos{\hspace{0.2mm}}

\makeatletter
\newcommand*\MY@rightharpoonupfill@{%
    \arrowfill@\relbar\relbar\rightharpoonup
}
\newcommand*\overrightharpoon{%
    \mathpalette{\overarrow@\MY@rightharpoonupfill@}%
}
\makeatother

\newlength{\depthofsumsign}
\setlength{\depthofsumsign}{\depthof{$\sum$}}
\newcommand{\nsum}[1][1.4]{
    \mathop{%
        \raisebox
            {-#1\depthofsumsign+1\depthofsumsign}
            {\scalebox
                {#1}
                {$\displaystyle\sum$}%
            }
    }
}


\begin{figure*}[!t]
    \centering
    \includegraphics[width=0.99\textwidth]{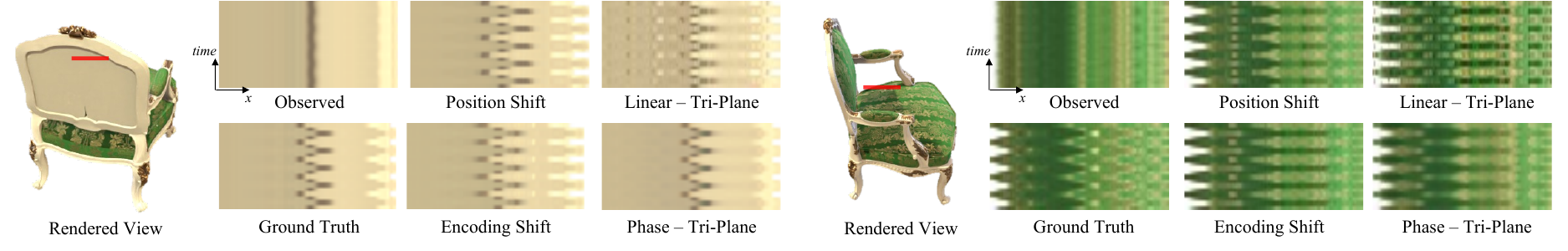}
    \includegraphics[width=0.99\textwidth]{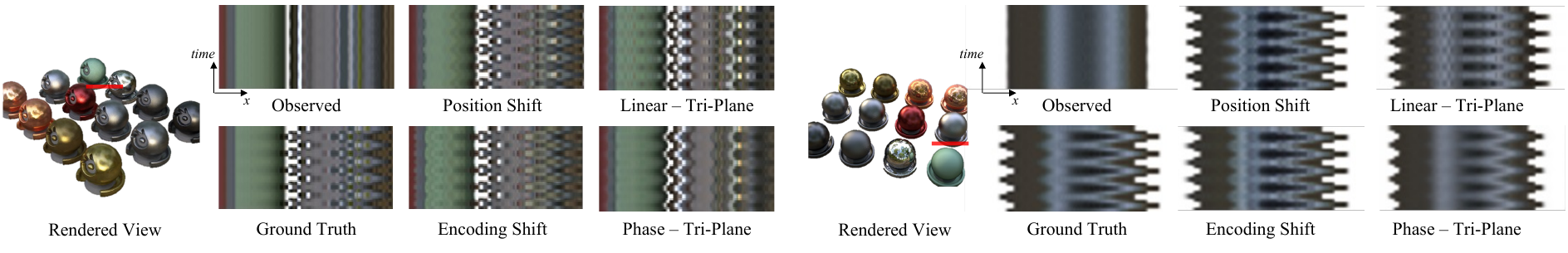}
\caption{\textbf{3D motion magnification on synthetic scenes.} 
We generated each synthetic scene by periodically vibrating object parts. 
We magnify the subtle motion encoded in NeRF reconstruction using the approaches discussed in Sec.~\ref{sec:method}, and visualize the motion here as a 2D space-time slice image. 
The corresponding location of each space-time slice is indicated by a red line on the rendered view. 
All four approaches successfully capture and magnify the motion, although the linear Eulerian approach, \textit{Linear - Tri-Plane}, is more prone to intensity overshooting~\cite{Wadhwa2013PhaseBased}, manifested as bright and dark spots in the space-time slice.}
    \label{fig:synthetic}
\end{figure*}

\section{Preliminaries}
\topic{Eulerian motion magnification.}
The Eulerian-based motion analysis focuses on the changes at a \emph{fixed} spatial location over time instead of tracking a specific particle (pixel).
The \emph{linear Eulerian approach} converts the color variation over time at each pixel into a 1D vector, using the Fourier transform to obtain its temporal frequency components, and filters the frequencies corresponding to the desired motion. 
The color intensity changes within the desired frequency range are then amplified and added back to the original values to create a motion-magnified video (where subtle motions become more visible).

To offer an intuition on why amplifying per-pixel color intensity could magnify motion across the frame, let $f(x, t) = g(x + \delta(t))$ denote a signal with motion over time described by the shift $\delta(t)$. 
The first-order Taylor series expansion of $g(x + \delta(t))$ about the point $x$ can be written as:
$$g(x) + g'(x)(x+\delta(t)-x) = g(x) + g'(x)\delta(t).$$
With observations at multiple timesteps $t$, we can easily filter out the static $g(x)$ term and keep the dynamic term $g'(x)\delta(t)$. 
If we multiply $g'(x)\delta(t)$ by $\alpha$ and add it back to the original signal, we get
$$g(x) + (1+\alpha)g'(x)\delta(t) \approx g(x + (1+\alpha)\delta(t)),$$
which is equivalent to magnifying the motion by $\alpha$.

The \emph{phase-based Eulerian approach} amplifies phase variations over time instead of color amplitude variations. 
The phase here is extracted from a complex steerable pyramid constructed from the original frames. 
The connection between phase and motion can be established through the Fourier shift theorem: if a function $f(x)$ is shifted by a distance $\delta$ in its domain, it would be equivalent to multiplying its Fourier component $\mathcal{F}(k)$ by a phase factor $e^{-i2\pi k\delta}$:
$$\mathcal{F}\{f(x-\delta)\}(k) = \mathcal{F}\{f(x)\}(k) e^{-i2\pi k\delta}, $$
where $\mathcal{F}$ denotes the Fourier transform operator and $k$ denotes the frequency component. 
In other words, extracting the phase changes over time reveals the motion-induced pixel shift in space by $\delta$. 
After amplifying the phase changes, the motion-magnified signal can be generated with an inverse Fourier transform.

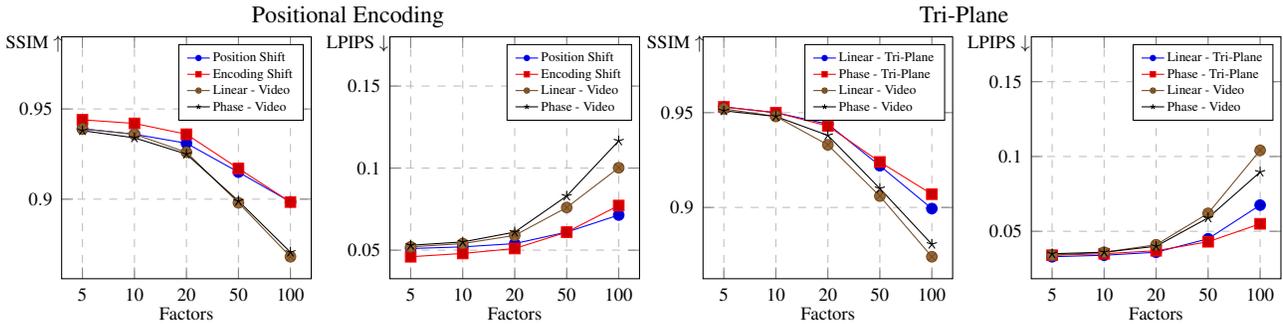
\begin{figure*}[h]
\begin{tabular}{m{0.17\textwidth}m{0.22\textwidth}m{0.24\textwidth}m{0.22\textwidth}m{0.22\textwidth}}
\centering
& \small{Positional Encoding} & & \small{Tri-Plane} &\\
\end{tabular}\vspace{-3pt} \\
\begin{tabular}{m{0.22\textwidth}m{0.22\textwidth}m{0.22\textwidth}m{0.22\textwidth}}
\centering
\begin{tikzpicture}
\begin{axis}[
x label style={at={(axis description cs:0.5,-0.08)},anchor=north}, y label style={at={(axis description cs:-0.08,.5)},anchor=south}, legend style={nodes={scale=0.5, transform shape}},
legend image post style={scale=0.5},
scaled ticks=false, 
tick label style={/pgf/number format/fixed,font=\scriptsize}, 
ymax = 0.99,
symbolic x coords={5,10,20,50,100},
xtick=data,
height=4.8cm,
width=4.9cm,
grid=major,
xlabel={\scriptsize Factors},
ylabel={\scriptsize SSIM $\uparrow$},
x label style={at={(axis description cs:0.5,-0.08)},anchor=north},
y label style={at={(axis description cs:-0.11,0.9)},rotate=270,anchor=south},
legend style={
cells={anchor=west},
legend pos=north east,
}
]

\addplot coordinates {(5,0.939) (10,0.936)  (20,0.931)  (50,0.915) (100, 0.8985) };

\addplot coordinates {(5,0.944) (10,0.942)  (20, 0.936)  (50,0.917) (100, 0.8984) };

\addplot coordinates {(5,0.939) (10,0.936)  (20,0.926)  (50,0.898) (100, 0.8681) };

\addplot coordinates {(5,0.938) (10,0.934)  (20,0.925)  (50,0.899) (100, 0.8706) };
\legend{Position Shift,Encoding Shift,Linear - Video,Phase - Video}
\end{axis}
\end{tikzpicture}
&
\begin{tikzpicture}
\begin{axis}[
legend style={nodes={scale=0.5, transform shape}},
legend image post style={scale=0.5},
scaled ticks=false, 
tick label style={/pgf/number format/fixed,font=\scriptsize}, 
ymax = 0.18,
symbolic x coords={5,10,20,50,100},
xtick=data,
height=4.8cm,
width=4.9cm,
grid=major,
xlabel={\scriptsize Factors},
ylabel={\scriptsize LPIPS $\downarrow$},
x label style={at={(axis description cs:0.5,-0.08)},anchor=north}, 
y label style={at={(axis description cs:-0.13,0.9)},rotate=270,anchor=south}, legend style={
cells={anchor=west},
legend pos=north east,
}
]
\addplot coordinates {
(5,0.051) (10, 0.052)  (20,0.054)  (50,0.061) (100, 0.0714)
};

\addplot coordinates {(5,0.046) (10,0.048)  (20,0.051)  (50,0.061)  (100, 0.0772)};

\addplot coordinates {(5,0.052) (10, 0.054)  (20, 0.059)  (50, 0.076)  (100, 0.1002)};

\addplot coordinates {(5,0.053) (10, 0.055)  (20,0.061)  (50,0.083)  (100, 0.1166)};
\legend{Position Shift,Encoding Shift,Linear - Video,Phase - Video}
\end{axis}
\end{tikzpicture}
& 
\begin{tikzpicture}
\begin{axis}[
x label style={at={(axis description cs:0.5,-0.08)},anchor=north}, y label style={at={(axis description cs:-0.08,.5)},anchor=south}, legend style={nodes={scale=0.5, transform shape}},
legend image post style={scale=0.5},
scaled ticks=false, 
tick label style={/pgf/number format/fixed,font=\scriptsize}, 
ymax = 0.99,
symbolic x coords={5,10,20,50,100},
xtick=data,
height=4.8cm,
width=4.9cm,
grid=major,
xlabel={\scriptsize Factors},
ylabel={\scriptsize SSIM $\uparrow$},
x label style={at={(axis description cs:0.5,-0.08)},anchor=north}, 
y label style={at={(axis description cs:-0.11,0.9)},rotate=270,anchor=south},
legend style={
cells={anchor=west},
legend pos=north east,
}
]
\addplot coordinates {
(5,0.953) (10, 0.950)  (20, 0.944)  (50, 0.922) (100, 0.8994)
};

\addplot coordinates {
(5,0.953) (10, 0.950)  (20,0.943)  (50,0.924) (100, 0.9070)
};

\addplot coordinates {
(5,0.952) (10, 0.948)  (20,0.933)  (50,0.906) (100, 0.8740)
};

\addplot coordinates {
(5,0.951) (10, 0.948)  (20,0.938)  (50,0.910) (100, 0.8806)
};

\legend{Linear - Tri-Plane,Phase - Tri-Plane,Linear - Video,Phase - Video}
\end{axis}
\end{tikzpicture}
&
\begin{tikzpicture}
\begin{axis}[
legend style={nodes={scale=0.5, transform shape}},
legend image post style={scale=0.5},
scaled ticks=false, 
tick label style={/pgf/number format/fixed,font=\scriptsize}, 
ymax = 0.18,
symbolic x coords={5,10,20,50,100},
xtick=data,
height=4.8cm,
width=4.9cm,
grid=major,
xlabel={\scriptsize Factors},
ylabel={\scriptsize LPIPS $\downarrow$},
x label style={at={(axis description cs:0.5,-0.08)},anchor=north}, 
y label style={at={(axis description cs:-0.13,0.9)},rotate=270,anchor=south},
 legend style={
cells={anchor=west},
legend pos=north east,
}
]
\addplot coordinates {
(5,0.033) (10, 0.034)  (20,0.036)  (50,0.045)  (100, 0.0675)
};

\addplot coordinates {
(5,0.034) (10,0.035)  (20,0.037)  (50,0.043)  (100, 0.0550)
};

\addplot coordinates {
(5,0.034) (10,0.036)  (20,0.041)  (50,0.062)  (100, 0.1041)
};
\addplot coordinates {
(5,0.035) (10,0.036)  (20,0.040)  (50,0.059)  (100, 0.0896)
};
\legend{Linear - Tri-Plane,Phase - Tri-Plane,Linear - Video,Phase - Video}
\end{axis}
\end{tikzpicture}
\end{tabular}
\caption{\textbf{Quantitative comparison.} 
We evaluate the quality of motion-magnified renderings as a function of the magnification factor used, using positional encoding (\textit{Left}) and tri-plane (\textit{Right}) as the point embedding function. 
With positional encoding, we evaluate two approaches to vary point embedding through phase shifts in the sine waves: 
\textit{Position Shift} (shifting each 3D point) and \textit{Encoding Shift} (shifting each frequency). 
With tri-plane, we evaluate two approaches to vary learned point embeddings: \textit{Linear - Tri-Plane} (linear magnification on tri-plane) and \textit{Phase - Tri-Plane} (phase-based magnification on tri-plane). 
For both embedding functions, we compare against two baseline methods for video motion magnification: \textit{Linear - Video} (linear magnification on the NeRF-rendered video) and \textit{Phase-Video} (phase-based magnification on the NeRF-rendered video). 
Results from two embedding functions are separated to enable better assessments of the impact of different magnification approaches and avoid confounding with the inherent performance gap between different embedding functions and MLP architectures.}
\label{fig:magfactor}
\end{figure*}

\topic{Neural radiance fields as 3D scene representations.}
NeRF models the radiance in a scene as a continuous function, which takes as input a 3D spatial coordinate ${p} \in \mathbb{R}^3$ and a viewing direction ${d} \in \mathbb{S}^2$, and outputs the radiance color $c$ (observed from viewing direction $d$) and density $\sigma$ at that point. 
Notably, the spatial coordinate $p$ is transformed into a feature representation through some embedding function ${E}$, before a projection function (MLP) regresses it into the final prediction:
$$f(p, d) = \text{MLP}({E}(p), d) = (c, \sigma).$$

With subtle and unknown scene motions, we assume this time-varying scene can be formulated as $f(p + \delta(p, t), d)$. 
If MLP is fixed across time, then the unknown motion $\delta(p, t)$ can be recovered by analyzing the temporal variations of ${E}(p, t)$.
However, where do we access ${E}(p, t)$? 
Whereas in the 2D video case, the data of interest is directly recorded by a camera and is available for analysis, here we only have access to a collection of 2D images that may have observed the 3D subtle motions during capture.
In the following subsection, we discuss how to reconfigure NeRF to model subtle 3D motions by varying the function ${E}(p, t)$.

\section{Method}
\label{sec:method}

We assume the availability of: 
1) Multi-view observations to reconstruct a static NeRF, and 
2) video recording of the subtle scene motions, either with a time-synchronized multi-camera setup or a single moving camera.

The general workflow of our method is as follows: 
1) We train a static NeRF from image observations that can be assumed as motionless. 
2) For each timestep $t \in [0, T-1]$ in the video observations, we finetune the embedding function $E_{t}$ so that the NeRF rendering matches with the observations at $t$. 
3) After finetuning all $T$ embedding functions $E_{t}$, we magnify motions by amplifying the temporal variations of each sampled point used in NeRF rendering.

In this section, we describe how we repurpose NeRF to capture subtle motions and perform magnification by analyzing the point embeddings learned by NeRF.
We begin our discussion with the base case of the standard NeRF with positional encoding as the point embedding function.
We then describe our preferred approach with tri-plane as the embedding function for NeRF, which leads to a natural integration with the phase-based Eulerian magnification technique previously designed for videos.

\subsection{NeRF with Positional Encoding}
We first describe how Eulerian magnification in the embedding space can be achieved on standard NeRF with positional encoding.
Motivated by prior work on motion-adjustable neural representations for video~\cite{Mai_2022_PhaseNIVR}, we keep the main backbone of NeRF intact and separately train a small MLP $g$ that learns to apply phase shifts in the positional encoding functions. 
However, different from prior work~\cite{Mai_2022_PhaseNIVR}, with video observations from the scene, we organize the images by their captured time and train a separate MLP $g$ for each timestep. 
Effectively, $g$ learns to adjust the embedded representation of each point so that the NeRF output from the projection function (MLP $\theta$) is consistent with the time-varying observations. 
Weights of the MLP $\theta$ are shared across all time steps, and the only difference lies in the point embeddings.
There are two options to induce phase shifts to the positional encoding function: position shift and encoding shift.

\begin{figure*}[!t]
    \centering
    \includegraphics[width=0.99\textwidth]{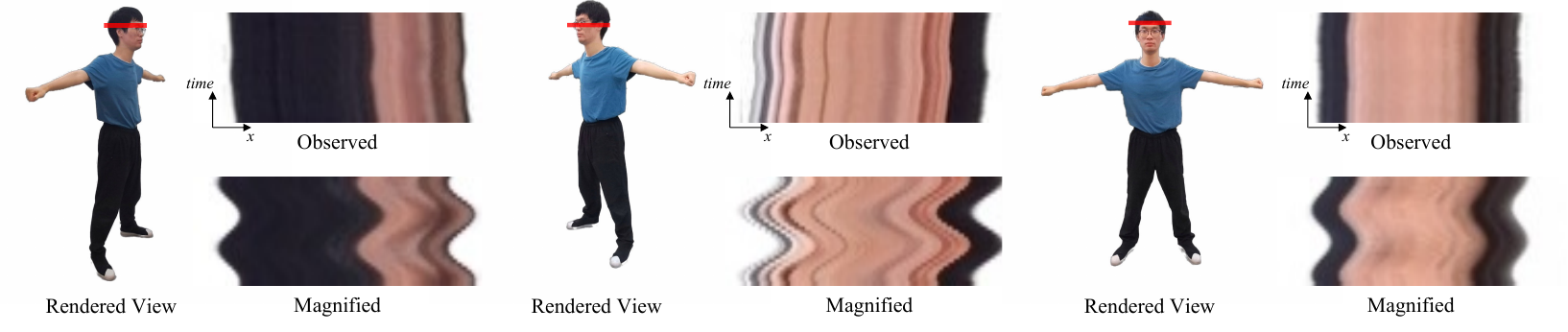}
\caption{\textbf{Real-world multi-view motion magnification}. 
Using multi-view videos from the HumanNerf dataset \cite{neural-human-radiance-field}, we can capture and magnify true 3D motion. 
We visualize the motion here as a 2D space-time image, where the corresponding location of each space-time slice is shown on the  rendered view as a red line.}
    \label{fig:humman}
\end{figure*}

\begin{figure*}[!t]
    \centering
    \includegraphics[width=0.99\textwidth]{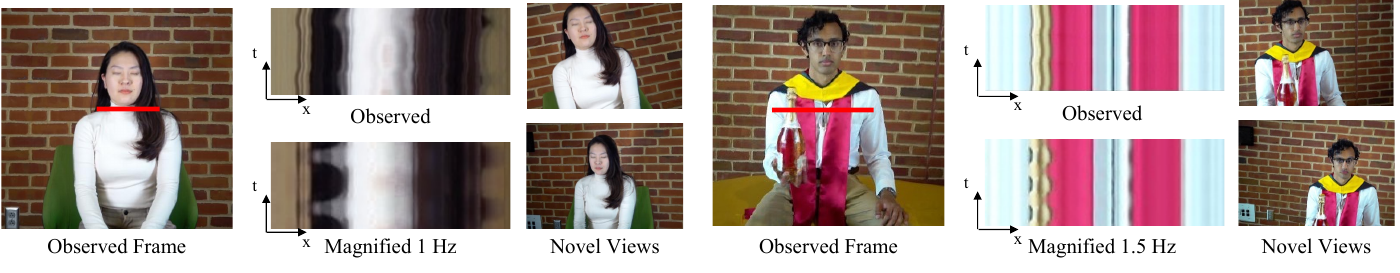}
\caption{\textbf{Real-world single-camera motion magnification}. 
Despite the monocular view of the dynamic scene, we can render novel views while magnifying subtle motion. We visualize the motion as a 2D image slice through time.
    }
    \label{fig:realresult}
\end{figure*}

\paragraph{Position shift.} 
To model motion exclusively through changing the point embeddings, we let $g$ directly \textbf{predict the 3D position shift of the queried point $p$}: $g(p, t) = \Delta{p} \in \mathbb{R}^{3}$. 
We add $\Delta{p}$ to $p$ before applying positional encoding, obtaining a time-varying point embedding function
\begin{align}
    E(p, t) = \text{PosEnc}(p + g(p, t)), \nonumber
\end{align}
which is equivalent to applying a phase shift of $\omega \cdot g(p, t)$ within each sine wave $\sin(\omega x)$ used in positional encoding. Note that the phase shifts for all $K$ frequencies have the same direction and only differ in magnitude, which is scaled by different $\omega \in [1, ..., K]$.

\paragraph{Encoding shift.} 
Note that motion does not only lead to geometric changes in the 3D scene since it would also cause appearance changes like shadows and reflections.
Therefore, attributing all the scene variation to shifts in 3D position is not sufficient.
Instead, we may let $g$ learn \textbf{a separate shift for each encoding frequency}. 
In other words, with $\phi_{\omega} \in \mathbb{R}^{3}$, let $g(p, t) = [\phi_{1}, \phi_{2}, ..., \phi_{K}] \in \mathbb{R}^{3\times K}$, and the point embedding function becomes
\begin{align}
    E(p, t) &= [\sin(\omega_{1} p + \phi_{1}), ..., \sin(\omega_{K}p + \phi_{K})]. \nonumber
\end{align}
This setup treats positional encoding as a feature generator that produces an embedding with $3K$ channels. 
As shown in later sections, this ``motion-agnostic" approach still learns to capture the true motion while outperforming the approach that only accounts for position shifts.

With either of these two approaches to vary the point embeddings in a standard NeRF with positional encoding, we can render magnified motions by linearly amplifying the temporal variations of $g(p, t)$, and then rendering the point's color and opacity with NeRF.

\subsection{NeRF with Tri-plane Learnable Embedding}
We now describe our preferred approach using tri-plane as the embedding function for NeRF. 
Later experiments suggest that the tri-plane-based approach achieved better magnification quality than the positional encoding approach.
Tri-plane~\cite{chan2022efficient,sfk_kplanes_2023} has been recently proposed as an efficient way to obtain learnable embedding for points in NeRF rendering.
Compared to NeRF with the analytical positional encoding, NeRF with tri-planes as the point embedding function can achieve similar representation capacity with far fewer MLP layers and, thus, faster inference.

In our use case of NeRF, the tri-plane formulation has a nice implication of reducing the 3D scene into a collection of 2D feature planes.
Such a decomposition preserves the relative spatial relationship between points, instead of randomly hashing points into features.
This observation suggests we may achieve 3D motion magnification by directly processing the feature planes and potentially outperforming the aforementioned linear magnification within the positional encoding function. 

Specifically, we train a separate tri-plane embedding function for each timestep while the MLP-based projection function is shared across time.
With this setup, all subtle temporal changes in the scene (motion or appearance) would need to be encapsulated in the temporal changes of the 2D feature images of the tri-plane. 
The point embedding function here can be written as $$E(p, t) = \text{Project}(p, \text{TriPlane}_{t}),$$
where the embedding of $p$ is obtained by projecting $p$ onto the tri-plane and aggregating the corresponding features.

With a separate tri-plane constructed for each timestep, we establish a key connection to prior video magnification methods: we essentially obtain a \textit{video} for each tri-plane feature channel, on which we could either apply \textbf{linear magnification} and amplify the temporal changes of each pixel in the feature image, or \textbf{phase-based magnification} with complex steerable pyramids constructed over each channel of the 2D feature image. 
Our experimental results confirm the feasibility of this 2D-inspired approach. 
The performance comparison in Section \ref{sec:synthetic_exp} between the linear and phase-based approaches also validates the original findings from when these two approaches were applied to perform Eulerian processing of color spaces.

\begin{figure}[t]
\begin{tabular}{m{0.20\textwidth}m{0.22\textwidth}}
\centering 
\begin{tikzpicture}
\begin{axis}[
legend style={nodes={scale=0.5, transform shape}},
legend image post style={scale=0.5},
scaled ticks=false, 
tick label style={/pgf/number format/fixed,font=\scriptsize, /pgf/number format/precision=2,/pgf/number format/fixed zerofill,}, 
ymax = 0.98,
symbolic x coords={5,10,15,20},
xtick=data,
height=4.8cm,
width=4.8cm,
grid=major,
xlabel={\scriptsize Angles},
ylabel={\scriptsize SSIM $\uparrow$},
x label style={at={(axis description cs:0.5,-0.08)},anchor=north}, 
y label style={at={(axis description cs:-0.15,1.03)},rotate=270,anchor=south},
legend style={
cells={anchor=west},
legend pos=north east,
}
]

\addplot coordinates {(5,0.961) (10,0.958)  (15,0.958)  (20,0.955)};

\addplot coordinates {(5,0.960) (10, 0.958)  (15,0.957)  (20,0.956)};

\addplot coordinates {(5,0.959) (10,0.957)  (15,0.956)  (20,0.954)};

\addplot coordinates {(5,0.958) (10,0.956)  (15,0.955)  (20,0.953)};

\legend{Linear - Tri-Plane,Phase - Tri-Plane,Linear - Video,Phase - Video}
\end{axis}
\end{tikzpicture}
&
\begin{tikzpicture}
\begin{axis}[
legend style={nodes={scale=0.5, transform shape}},
legend image post style={scale=0.5},
scaled ticks=false, 
tick label style={/pgf/number format/fixed,font=\scriptsize, /pgf/number format/precision=2,/pgf/number format/fixed zerofill,}, 
ymax = 0.04,
symbolic x coords={5,10,15,20},
xtick=data,
height=4.8cm,
width=4.8cm,
grid=major,
xlabel={\scriptsize Angles},
ylabel={\scriptsize LPIPS $\downarrow$},
x label style={at={(axis description cs:0.5,-0.08)},anchor=north}, 
y label style={at={(axis description cs:-0.15,1.03)},rotate=270,anchor=south},
legend style={
cells={anchor=west},
legend pos=north east,
}
]

\addplot coordinates {(5,0.0284) (10,0.0291)  (15,0.0297)  (20,0.0317)};

\addplot coordinates {(5,0.0287) (10, 0.0294)  (15,0.0302)  (20,0.0304)};

\addplot coordinates {(5,0.0304) (10,0.0307)  (15,0.0315)  (20,0.0314)};

\addplot coordinates {(5,0.0305) (10,0.0309)  (15,0.0315)  (20,0.0320)};

\legend{Linear - Tri-Plane,Phase - Tri-Plane,Linear - Video,Phase - Video}
\end{axis}
\end{tikzpicture}
\end{tabular}

\caption{\textbf{Varying the angles of deviation from observed views}. As the deviation angle increases, magnifying through the embedding space consistently outperforms the baseline approaches that operate on the color space.
}
\label{fig:angledeviation}
\end{figure}
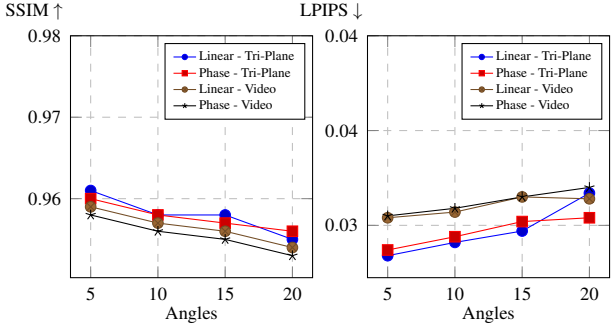 
\begin{figure}[]
\begin{tabular}{m{0.20\textwidth}m{0.22\textwidth}}
\centering
\begin{tikzpicture}
\begin{axis}[
legend style={nodes={scale=0.5, transform shape}},
legend image post style={scale=0.5},
scaled ticks=false, 
tick label style={/pgf/number format/fixed,font=\scriptsize, /pgf/number format/precision=2,/pgf/number format/fixed zerofill,}, 
ymax = 0.97,
symbolic x coords={0.01,0.05,0.1,0.2},
xtick=data,
height=4.8cm,
width=4.8cm,
grid=major,
xlabel={\scriptsize $\sigma^{2}$},
ylabel={\scriptsize SSIM $\uparrow$},
x label style={at={(axis description cs:0.5,-0.08)},anchor=north}, 
y label style={at={(axis description cs:-0.11,1.03)},rotate=270,anchor=south},
legend style={
cells={anchor=west},
legend pos=north west,
}
]

\addplot coordinates {(0.01,0.8366) (0.05,0.8323)  (0.1,0.8336)  (0.2,0.8066)};

\addplot coordinates {(0.01,0.8631) (0.05,0.8573)  (0.1,0.8621)  (0.2,0.8478)};

\addplot coordinates {(0.01,0.8255) (0.05,0.8193)  (0.1,0.8224)  (0.2,0.8116)};

\addplot coordinates {(0.01,0.8388) (0.05,0.8362)  (0.1,0.8459)  (0.2,0.8388)};

\legend{Linear - Tri-Plane,Phase - Tri-Plane,Linear - Video,Phase - Video}
\end{axis}
\end{tikzpicture}
&
\begin{tikzpicture}
\begin{axis}[
legend style={nodes={scale=0.5, transform shape}},
legend image post style={scale=0.5},
scaled ticks=false, 
tick label style={/pgf/number format/fixed,font=\scriptsize, /pgf/number format/precision=2,/pgf/number format/fixed zerofill,}, 
ymax = 0.2,
symbolic x coords={0.01,0.05,0.1,0.2},
xtick=data,
height=4.8cm,
width=4.8cm,
grid=major,
xlabel={\scriptsize $\sigma^{2}$},
ylabel={\scriptsize LPIPS $\downarrow$},
x label style={at={(axis description cs:0.5,-0.08)},anchor=north}, 
y label style={at={(axis description cs:-0.11,1.03)},rotate=270,anchor=south},
legend style={
cells={anchor=west},
legend pos=north west,
}
]

\addplot coordinates {(0.01, 0.1420) (0.05, 0.1473)  (0.1, 0.1433)  (0.2, 0.1959)};

\addplot coordinates {(0.01, 0.1174) (0.05, 0.1242)  (0.1, 0.1162)  (0.2, 0.1397)};

\addplot coordinates {(0.01, 0.1324) (0.05, 0.1395)  (0.1, 0.1358)  (0.2, 0.1478)};

\addplot coordinates {(0.01, 0.1341) (0.05, 0.1405)  (0.1, 0.1257)  (0.2, 0.1414)};

\legend{Linear - Tri-Plane,Phase - Tri-Plane,Linear - Video,Phase - Video}
\end{axis}
\end{tikzpicture}
\end{tabular}

\caption{\textbf{Varying noise levels in training views}. When we perform Eulerian magnification on the tri-plane embedding space, \textit{Phase - Tri-Plane} is more robust than \textit{Linear - Tri-Plane} in the presence of noise. 
The finding is analogous to previous analysis~\cite{Wadhwa2013PhaseBased} on color space magnification (validated by the two baseline results: \textit{Linear - Video} and \textit{Phase - Video}).}
\label{fig:noiselevel}
\end{figure}
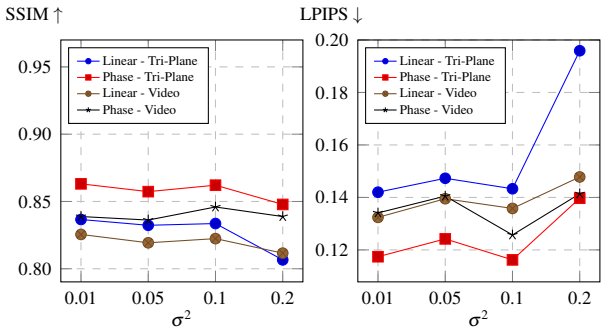

\section{Experimental Results}
\label{sec:result}
We evaluate the performance of the proposed method using synthetic scenes in Section \ref{sec:synthetic_exp}. 
We create ground truth sequences for the magnified motion and quantitatively compare the different approaches for 3D motion magnification.
We then present our results on real-world captured data in Section \ref{sec:real_exp}. 
We first deploy our method on real-world multi-view video observations. 
After validating its effectiveness on real-world multi-view data, we further apply our method to real-world video sequences captured using a \textit{single-camera}. 
As a result of extending motion magnification to 3D, our method successfully magnifies 3D motions from handheld-captured videos with camera shake, a scenario unattainable by prior work that focused on stabilized 2D videos. 

\subsection{Implementation Details}
\paragraph{Positional encoding as point embedding.} 
We train a 3-layer MLP with 32 hidden channels to predict $g(p, t)$ and apply the resulting phase shifts in positional encoding. We implement the network with nerfacc~\cite{li2022nerfacc}. To train a static NeRF for the first timestep, we optimize for 50,000 steps, and for the remaining timesteps, we finetune the embedding function for 10,000 steps. 
After training, for each point $p$, we obtain its time-varying phase shifts $g(p, t)$. 
To render with magnified motion within a time window $[0, ..., T-1]$, we Fourier transform $[g(p, 0), ..., g(p, T-1)]$ along the time dimension, use a bandpass filter to isolate the motions within the frequency range of interest, amplify the components within the passband range, and then apply inverse Fourier transform to obtain the magnified predictions. 
The magnified predictions are added inside the positional encoding as phase shifts, followed by the standard NeRF rendering with MLP inference.

\paragraph{Tri-plane as point embedding.} 
Our implementation builds on K-Planes~\cite{sfk_kplanes_2023}. 
To aggregate the embedding from different planes, we adopt concatenation instead of the default Hadamard product, and we only set the triplanes at a single scale for simplicity. 
To train a static NeRF for the first timestep, we optimize for 30,000 steps, and for the remaining timesteps we finetune the embedding function for 10,000 steps. 
After training, we compose a video for each feature plane within $[0, ..., T-1]$. 
Then, we apply 2D magnification methods directly on these \emph{feature videos}. 
For the \textit{linear} approach, we temporally filter and amplify the feature value variations within a frequency range. 
For the \textit{phase-based} approach, we construct a complex steerable pyramid over the feature image, temporally filter and amplify the phase variations, and then collapse the pyramid back into the image space; the resulting feature image would exhibit magnified motions~\cite{Wadhwa2013PhaseBased}. To produce motion-magnified rendering, we embed each sample point with the processed triplane features and then use the MLP to project the embedding into color and density output as usual.

\subsection{Synthetic Scenes}
\label{sec:synthetic_exp}

\topic{Data Generation.} We use the standard Blender scenes \cite{mildenhall2020nerf} and simulate motions in different object parts. We render each scene for one second at 30 frames per second. 
The simulated motions are periodic, ranging from 3 Hz to 5 Hz. 
We also render sequences with ground truth magnified motions under factors 5, 10, 20, 50, and 100.

\topic{Qualitative Evaluation.} We magnify motions in the Blender scenes with the approaches described in Sec.~\ref{sec:method}.
Results in Figure~\ref{fig:synthetic} show that all four approaches successfully model and magnify tiny motions in the observations. 
Consistent with previous findings in video magnification~\cite{Wadhwa2013PhaseBased}, the linear Eulerian approach leads to more artifacts, such as clipped color intensities and noise amplification.

\topic{Quantitative Evaluation.} We evaluate the results against ground truth magnified frames using structural similarity index measure (SSIM) \cite{ssim}, and LPIPS \cite{zhang2018perceptual} with AlexNet \cite{alexnet} as the backbone. 
We also render a video from the trained NeRF without magnification for baseline comparisons at each test view. 
We then apply classical 2D methods directly on the video to magnify subtle motions. 
As shown in Figure~\ref{fig:magfactor}, our 3D magnification methods outperform the 2D baseline methods in producing motion-magnified renderings consistent with the ground truth renderings.

\topic{Sensitivity Aanalysis.}
In Figure~\ref{fig:angledeviation}, we analyze magnification quality as the test viewpoints deviate from the observed viewpoints. 
In Figure~\ref{fig:noiselevel}, we plot magnification quality as noise levels increase in the captured frames. 
Previous work~\cite{Wadhwa2013PhaseBased} on color space magnification has found that the phase-based approach is less sensitive to noise than the linear approach; we observe similar phenomenon during our magnification in the embedding space.

\subsection{Real-world Scenes}
\label{sec:real_exp}

We test our methods on several real-world scenes captured with different camera setups.

\topic{Multi-Camera Setup.}
We first validate our method on the publicly available dataset from HumanNerf~\cite{Zhao_2022_CVPR}, comprising short videos from six cameras simultaneously capturing a scene with a person standing in the center. 
We extract a brief period where the person is relatively static but still exhibits subtle body movements.
In Figure \ref{fig:humman}, we present the magnified results from different viewpoints; the full videos are available in the supplementary material.

\topic{Single-Camera Setup.}
As a multi-camera setup may be prohibitively inconvenient and expensive for many users, we further deploy our method on a single-camera setup. 
We design a capture procedure that consists of two stages. 
Step 1: we first capture a moving camera video of the static scene, which will be used to train a static NeRF. 
Step 2: we capture a single-view video of the dynamic scene, which is used to finetune the point embedding function in NeRF to model the time-varying scene.
After the two-stage capture, we perform NeRF training and render magnified motions using the previously described pipeline. 
This two-step capture approach is prevalent in video-based motion magnification application scenarios, where people identify unwanted motion in static civil structures under normal conditions~\cite{shang2018multi_civil, lado2023learning_civil}. 
We also highlight that our method \textbf{supports handheld capture and does not require a tripod}, unlike previous 2D-based approaches that would fail without a steady capture, as shown in Figure \ref{fig:handheld}. 
This is possible since our method uses the estimated poses of each frame independently when updating the radiance fields on the dynamic sequence. 
Hence, an accurate pose estimation removes the need for a perfectly still capture.
We show the results using monocular capture in Figure \ref{fig:teaser} of a gymnast holding a handstand and in Figure \ref{fig:realresult} on a person trying to balance on one leg and a person breathing. 
We also show observed space-time slices compared to our magnified ones with NeRF reconstruction.

\section{Limitations}
\label{sec:limitations}
Data captured in real-world environments can be blurry due to defocus and camera shake, degrading the quality of NeRF. 
The training of NeRF assumes the knowledge of camera poses, so its performance depends heavily on the accuracy of pose estimation, often using RGB-based structure-from-motion (SfM) algorithms \cite{schoenberger2016sfm, schoenberger2016mvs}. 
The estimations are mostly reliable but not flawless due to lens distortions that require specific calibrations, which may not be performed by common pipelines for NeRF-based 3D reconstruction.
More importantly, inaccurate pose estimation would exacerbate the ambiguity between \textit{camera motion} and \textit{scene motion}, which could hinder magnifying subtle scene motions or lead to false motion magnification.
Therefore, real-world data should be captured under conditions where accurate camera intrinsic and extrinsic parameters are accessible, either from {\it reliable} RGB-based SfM with textured surfaces in the scene, or from cameras that support 6-DoF tracking during capture.
\section{Conclusion}
\label{sec:conclusions}
We present a 3D motion magnification method that applies Eulerain processing principles to analyzing NeRF embeddings over time. 
While classical magnification methods (developed originally for 2D videos) process pixel colors directly, we show that processing the point embeddings of NeRF successfully generalizes those approaches and allows magnifying motions in 3D renderings.
We believe our work will motivate further research towards integrating traditional signal processing techniques into neural fields.

\clearpage

{\small
\bibliographystyle{ieee_fullname}
\bibliography{egbib}
}

\end{document}